\documentclass{article}
\usepackage{ijcai25}

\usepackage{float}
\usepackage{multirow} 
\usepackage{amsfonts}
\usepackage{arydshln}
\usepackage{subfigure}
\usepackage{comment}

\usepackage{times}
\usepackage{soul}
\usepackage{url}
\usepackage[hidelinks]{hyperref}
\usepackage[utf8]{inputenc}
\usepackage[small]{caption}
\usepackage{graphicx}
\usepackage{amsmath}
\usepackage{amsthm}
\usepackage{booktabs}
\usepackage{algorithm}
\usepackage{algorithmic}
\usepackage[switch]{lineno}

\title{Instance Relation Learning Network with Label Knowledge Propagation for Few-shot Multi-label Intent Detection}

\author{
Shiman Zhao$^{1,2,3}$
\and
Shangyuan Li$^{1,2,3}$ 
\and 
Wei Chen\thanks{Corresponding author}$^{1,2,3}$ 
\and 
Tengjiao Wang$^{1,2,3,4}$ \\
Jiahui Yao$^{1,2,3}$ 
\and
Jiabin Zheng$^{1,2,3}$ 
\And
Kam Fai Wong$^{1,4}$ 
\affiliations
$^1$Key Lab of High Confidence Software Technologies (MOE), School of Computer Science, Peking University, Beijing, China. \\
$^2$Research Center for Computational Social Science, Peking University, Beijing, China. \\
$^3$Institute of Computational Social Science, Peking University, Qingdao, China. \\
$^4$Department of Systems Engineering and Engineering Management, The Chinese University of Hong Kong, Hong Kong, China.
\emails
\emails
shiman.zha@gmail.com, \{pekingchenwei, tjwang, isssyaojh,jiabinzheng\}@pku.edu.cn
}

\begin{document}

\maketitle

\begin{abstract}
Few-shot Multi-label Intent Detection (MID) is crucial for dialogue systems, aiming to detect multiple intents of utterances in low-resource dialogue domains.
Previous studies focus on a two-stage pipeline.
They first learn representations of utterances with multiple labels and then use a threshold-based strategy to identify multi-label results.
However, these methods rely on representation classification and ignore instance relations, leading to error propagation.
To solve the above issues, we propose a multi-label joint learning method for few-shot MID in an end-to-end manner, which constructs an instance relation learning network with label knowledge propagation to eliminate error propagation.
Concretely, we learn the interaction relations between instances with class information to propagate label knowledge between a few labeled (support set) and unlabeled (query set) instances.
With label knowledge propagation, the relation strength between instances directly indicates whether two utterances belong to the same intent for multi-label prediction.
Besides, a dual relation-enhanced loss is developed to optimize support- and query-level relation strength to improve performance.
Experiments show that we outperform strong baselines by an average of 
9.54\% AUC and 11.19\% Macro-F1 in 1-shot scenarios.
\end{abstract}

\section{Introduction}

Multi-label Intent Detection (MID) \cite{zhang2021few,pham2023misca,tu2023joint} is a crucial component in dialogue systems \cite{zhang2020task,han2021fine}, which aims to identify multiple user intents in a given utterance. 
For example, in the sentence ``\textit{What time is my meeting and what day?}'', the MID task could detect two user intents,  ``\textit{request\_time}'' and ``\textit{request\_day}''.
With conversational artificial intelligence, MID has attracted widespread attention in academia and industry. 
Previous methods heavily rely on large amounts of labeled data, but their performances drop significantly in low-resource dialogue domains.
Few-shot MID is indispensable because it can detect multiple intents of utterances when only a few labeled data are provided.

\begin{figure}[t]
 \centering
 \includegraphics[width=3.4 in, height=1.8 in]{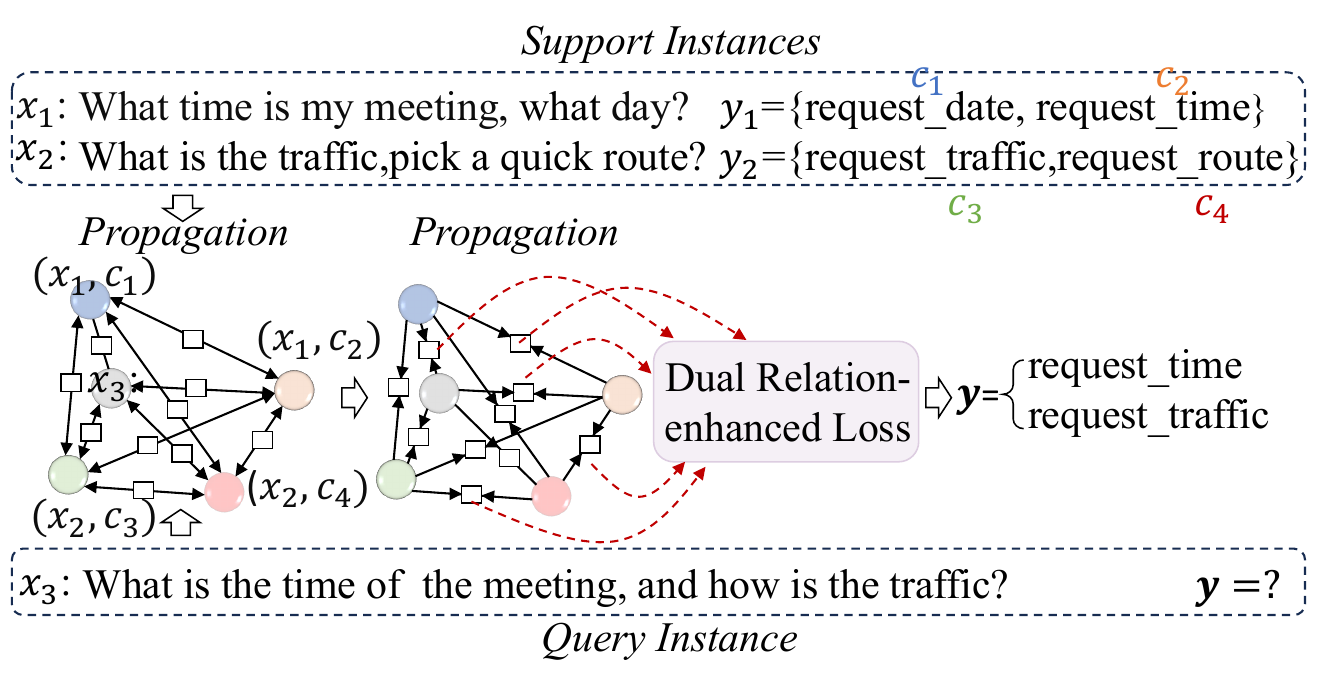}
 \caption{An example of the 4-way 1-shot setting.} 
 \label{example}  
\end{figure}

Existing methods \cite{HouDetection} mainly focus on a two-stage pipeline: (1) utterance representation learning and (2) multi-label results identification.
Specifically, they first learn different representations for an utterance with multiple intents and then use these representations to obtain multi-label results through a threshold-based strategy.
However, these methods heavily rely on the representation classification of utterances, leading to error propagation easily. 
For example, in Figure~\ref{example}, 
they are difficult to obtain well-separated representations for intents ``\textit{request\_time}'' and ``\textit{request\_date}'' because the utterance representations of these two intents could be closer to each other due to originating from the same utterance and expressing similar label semantics, as well as ``\textit{request\_traffic}'' and ``\textit{request\_route}''. 
Affected by multi-label noise, they may fail to obtain reliable representations, leading to low performance.
Although DCKPN \cite{zhang2023dual} learns utterance representations with class information to alleviate this issue, it still underperforms in this case due to error propagation.


To solve the above issues, we are the first to propose a multi-label joint learning method for few-shot MID, which designs an instance relation learning network with label knowledge propagation to directly gain multi-label results in an end-to-end manner, addressing error propagation. 
Concretely, the proposed method explicitly models the relationship between intra- and inter-class instances and propagates label knowledge between instances to capture strong interaction.
With label knowledge propagation, the proposed method could use the relation strength between an instance pair to indicate whether these two instances belong to the same label well for multi-label inference, eliminating the negative effect caused by error propagation.
Besides, a dual relation-enhanced loss is designed to enforce support-level and query-level relation strength to further improve performance.
The support-level loss enhances the relation strength between instances with the same label connected to the support instance and weakens the relation strength between those with different labels to promote label knowledge propagation.
The query-level loss encourages the maximization of relation strength between a query instance and its multiple relevant support instances while minimizing relation strength between the query and its irrelevant support instances to promote multi-label prediction.
Extensive experiments show that our proposed method performs well, especially an average improvement of 11.19\% Macro-F1 score in the 1-shot setting.
The contributions are summarized as follows:

\begin{itemize}
\item We propose a multi-label joint learning method for few-shot MID, which constructs an instance relation learning network with label knowledge propagation to guide query inference explicitly. The method eliminates error propagation and works well in low-resource domains.

\item We model the relationship between intra- and inter-instances to directly indicate whether two utterances belong to the same intent for multi-label prediction instead of depending on representation classification.

\item We design a dual relation-enhanced loss to optimize the support- and query-level interaction relations between instances and strengthen label knowledge propagation to further improve performance. 

\item Extensive experiments show that our proposed method outperforms strong baselines and obtains significant performance on the few-shot MID task. 

\end{itemize}

\section{Related Work}

\subsection{Few-shot Learning}

Few-shot learning has achieved great progress in Computer Vision (CV) \cite{ouyang2022self,chen2025dequantified} and Natural Language Processing (NLP) \cite{li-etal-2023-codeie,chan-etal-2023-shot,zhou2025valuing}.
Meta-learning is the mainstream method \cite{zhang2022protgnn,liang2023few,zhou2024boosting,zhou2024adversarial} for few-shot learning, and it includes model-based~\cite{TsendsurenMeta}, optimization-based~\cite{lee2019meta}, and metric-based~\cite{wang2021neural,lv2021learning,zhao2022label,chen-etal-2023-consistent} methods.  
Among these methods, metric-based methods are the most popular and potential research works. 
However, few-shot learning methods mostly focus on single-label prediction \cite{zhang2022fine,yehudai2023qaid,du2023all}, where they assign an intent to each utterance.
Obviously, these methods become problematic in real-world scenarios because an utterance may have multiple intents.
Few-shot MID could be regarded as a multi-label classification task.
Previous few-shot multi-label classifications focus on CV domain \cite{yan2022inferring}, audio domain \cite{8901732}, and sentiment analysis \cite{hu2021multi,Zhao2023learning}.
Few-shot MID is still in its infancy, and a few methods are available.

\subsection{Few-shot Multi-label Intent Detection}
Existing methods mostly focus on a two-stage pipeline. 
Hou et al. \shortcite{HouDetection} learn instance representations for multiple labels in the first stage. The second stage computes the relevance scores between the learned representations and different labels. Then, it uses a meta-calibrated dynamic threshold to perform multi-label classification based on relevance score ranking.
However, they ignore the interaction relations between instances and do not handle the identical representation of instances with multiple labels.
Graph Neural Networks (GNNs) \cite{MetricFree,sun2023self,chen2022brainnet} have shown great progress in modeling the relationship between instances using graph structure.
Zhang et al. \shortcite{zhang2023dual} propose a dual-class knowledge propagation network, which uses an instance-level GNN and a class-level GNN to learn instance representations. Then, they use a label count estimation to predict the label number of each instance.
However, these methods rely on the representation classification in a pipeline, inevitably causing error propagation.

Different from the aforementioned methods, our proposed method is the first to propose multi-label joint learning for few-shot MID in an end-to-end manner. 
We design an instance relation learning network with label knowledge propagation to explicitly explore the interaction relations between instances, which eliminates the negative effect caused by error propagation and works well in low-resource domains.

\begin{figure*}[t]
 \centering
 \includegraphics[width=6.8 in, height=3.8 in]{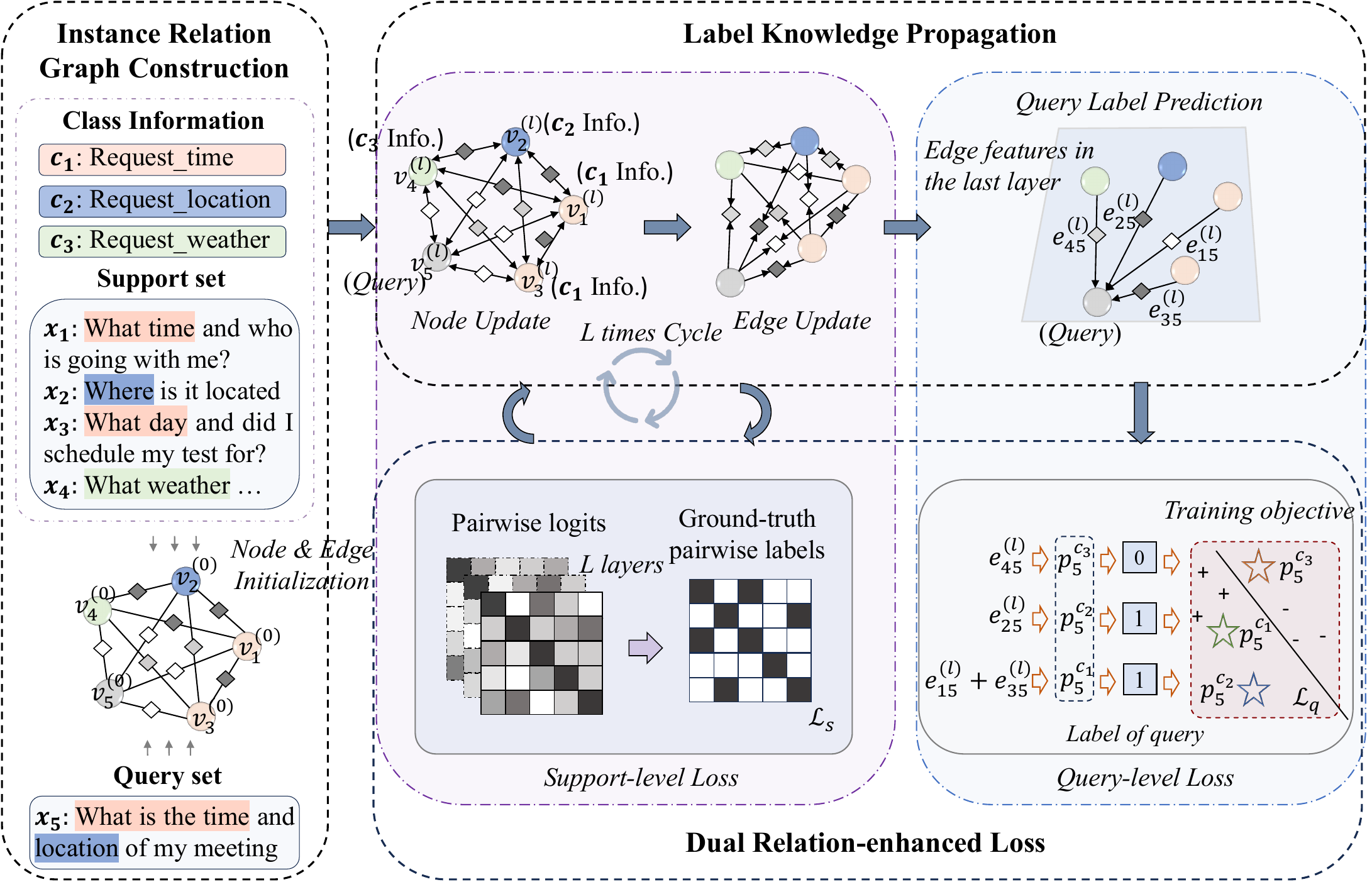}
 \caption{The overall architecture of the proposed method.} 
 \label{framework}  
\end{figure*} 

\section{Methodology}

\subsection{Task Formulation} \label{PF}
We formulate the few-shot MID task in meta-learning, which is commonly employed in current studies \cite{yu2022hybrid,fang2023manner,ma2023coarse} to generalize prior knowledge to a low-resource target domain from resource-rich source domains.
Our proposed method trains the model in a source domain ${\mathcal{D}_{source}}$ by building a collection of meta-tasks and tests it in a target domain ${\mathcal{D}_{target}}$ with other meta-tasks. 
Each meta-task consists of a support set and a query set.
Specifically, the support set $\mathcal{S}=\{({{x}}_i, {y}_i)\}_{i=1}^{N \times K}$ includes $N$ classes ($N$-way), and each class has $K$ instances ($K$-shot).
As a multi-label task, few-shot MID could assign more than one label to each instance. Briefly, for an instance ${x}_i$, its corresponding label is ${y}_i=\{y_i^1,y_i^2,...,y_i^N\}\in \{1,0\}^{N}$, where $y_i^{c_k}=1$ represents ${x_i}$ belongs to class $c_k$.
Besides, $\mathcal{Q}=\{({x}_j)\}_{j=1}^{T}$ is called the query set, and $T$ is the number of query instances.
The training objective is to minimize the query loss of $\mathcal{Q}$ based on $\mathcal{S}$ in ${\mathcal{D}_{source}}$. 
In the testing phase, we construct $\mathcal{S}$ and $\mathcal{Q}$ in ${\mathcal{D}_{target}}$.
Therefore, the proposed method predicts the query labels of $\mathcal{Q}$ based on $\mathcal{S}$ to verify the effectiveness of the model in the target domain ${\mathcal{D}_{target}}$.

\subsection{Overall Framework}

This section describes the proposed method for few-shot MID, as illustrated in Figure \ref{framework}.
We use the support set and the query set to construct a fully connected instance relation graph.
 Each node is an utterance, and each edge represents the relation strength between two connected nodes.  
In the graph network, our proposed method explicitly models the relationship between intra- and inter-class instances, which propagates label knowledge from support instances to query instances. Besides, a dual relation-enhanced loss is introduced to promote inference by enhancing support-level relation strength and query-level relation strength.

\subsection{Instance Feature Extraction} \label{FE}

Given an utterance $x$ with $n$ words, the utterance is defined as ${x} = \{ {{{w}}_1},{w_2},...,{w_n}\} $. We use a pre-trained text encoder like BERT \cite{devlin2019bert} to encode each word and generate hidden states ${H}^{x} = [h_1,h_2,...,h_n] \in \mathbb{R}^{n \times d}$. $d$ is the dimension of hidden states.
Besides, the corresponding class description ${c}$ with $m$ words is encoded into hidden states ${H}^{c} = [h_1,h_2,...,h_m] \in \mathbb{R}^{m \times d}$. 

\subsubsection{Support Instance Feature}
As an utterance has multiple intents (a.k.a. labels), it could be classified into different classes.
Support utterances for each class could express better  class-related intent semantics.
To extract class-related semantic features, we concatenate a support utterance and its class text ${H}^s = [{H}^{x}:{H}^{c}] \in \mathbb{R}^{(n+m) \times d}$ and utilize a self-attention mechanism \cite{Yan2020,liu2022label,du2022flow} to learn attention weights $A$ for each word, which generates the final representation of the support utterance. Specifically,

\begin{gather}
A=\text{softmax}(W_{2} \tanh(W_{1}H^s + b_1)+b_2), 
\label{FE_s1}
\end{gather}
\begin{gather}
{u}_s = {W_3}\text{ReLU}(H^sA^T)+b_3,
\label{FE_s2}
\end{gather}
where ${u}_s\in \mathbb{R}^{d}$ is the final feature representation of the support utterance. $W_1$, $W_2$, $W_3$, $b_1$, $b_2$ and $b_3$ are trainable parameters. 

\subsubsection{Query Instance Feature} 
For each query utterance, a mean pooling layer \cite{ChenExtracting} is utilized to calculate the average of hidden states across all words, serving as the feature representation of the query utterance.
\begin{equation}
{u}_q = \text{MeanPooling}(H^x).
\end{equation}
where ${u}_q\in \mathbb{R}^{d}$ is the query feature.

\subsection{Instance Relation Graph Construction}

Graph structure is beneficial for exploring the interaction relations between instances. Therefore, the proposed method utilizes the graph network to model the relationship between instances. 
As illustrated in Figure \ref{framework}, the instance relation graph is denoted as $\mathcal{G}=(\mathcal{V},\mathcal{E})$, where $\mathcal{V}$ is the node set and $\mathcal{E}$ is the edge set. 
The nodes in $\mathcal{V}$ correspond to instance features, and the edge $e_{ij} \in \mathcal{E}$ encodes the relationship between the $i^{th}$ instance and the $j^{th}$ instance.

\subsubsection{Node Initialization} 
The nodes in the graph $\mathcal{G}$ are defined as $\mathcal{V} = \{v_i\}_{i=1}^M$, where $M = |\mathcal{S}|+|\mathcal{Q}|$, and $v_i$ is the feature representation of the $i^{th}$ instance in the graph network. 
Nodes are initialized as $v_i^{(0)}=f_{emb}(x_i)$, where $f_{emb}(*)$ is the feature extraction function.

\subsubsection{Edge Initialization} 
The edges in the graph $\mathcal{G}$ are defined as
$\mathcal{E}=\{e_{ij}\}_{i,j}^{M}$ and represent the relationship between nodes.
Besides, $e_{ij}$ expresses the relation strength between two connected nodes and could be regarded as a probability that $v_i$ and $v_j$ are from the same class, reflecting intra-class similarity and inter-class dissimilarity.
To better model the relationship between instances, the instance features are projected as key and query \cite{VSInstance}. For the $i^{th}$ instance, its key ($k_{i}$), query ($q_i$) and node pairwise logits ($e_{ij}$) are as follows:
\begin{gather}
k_i^{(0)}=W_k^{(0)} \cdot v_i^{(0)}, 
\end{gather}
\begin{gather}
q_i^{(0)}=W_q^{(0)} \cdot v_i^{(0)},
\end{gather}
\begin{gather}
r_{ij}^{(0)} = q_i^{(0)}(k_j^{(0)})^T,
\label{edge1}
\end{gather}
where $W_{k}^{(0)}$ and $W_{q}^{(0)}$ are trainable parameters. The edge features are written as follows:

\begin{equation}
e_{ij}^{(0)} = \left\{\begin{matrix} 
   r_{ij}^{(0)} , &\ if \ \{v_i^{(0)}, \ v_j^{(0)}\} \in \mathcal{S}\\
 0 , &\ otherwise\\
\end{matrix}\right.
\end{equation}
where $e_{ij}^{(0)}$ represents the edge feature between nodes $v_i$ and $v_j$ in the $0^{th}$ layer.

\subsection{Label Knowledge Propagation}
The proposed method incorporates class information into representations of support instances in Equation \ref{FE_s1} and Equation~\ref{FE_s2}.
With message passing, the proposed method propagates label knowledge between support and query instances to enrich node and edge features.

\subsubsection{Node Update}
The proposed method propagates label knowledge from neighbors to enrich node features. 

\begin{equation}
\tilde{v}_i^{(l)} = \sum_{j}\frac{e_{ij}^{(l-1)}}{\sum_{k}e_{ik}^{(l-1)}}v_{j}^{(l-1)},
\end{equation}
\begin{equation}
\resizebox{0.86\hsize}{!}{$\begin{aligned}
{v}_i^{(l)} & = W_4 \text{ReLU}(\text{MLP}(\tilde{v}_i^{(l)})+{v}_i^{(l-1)})+b_4,
\end{aligned}$}
\end{equation}
where ${v}_i^{(l)}$ is the $i^{th}$ node feature in the $l^{th}$ layer. $\text{MLP}(*)$ is a multi-layer perceptron. $W_4$ and $b_4$ are trainable parameters. 

\subsubsection{Edge Update} 
With the graph network, the proposed method propagates label knowledge to adjust edge features from the latest node features.
The edge features are written as follows:

\begin{gather}
k_{i}^{(l)}=W_{k}^{(l)} \cdot v_i^{(l)}, 
\end{gather}
\begin{gather}
q_i^{(l)}=W_q^{(l)} \cdot v_i^{(l)},
\end{gather}
\begin{gather}
e_{ij}^{(l)} = q_i^{(l)}(k_j^{(l)})^T,
\label{edge2}
\end{gather}
where $W_{k}^{(l)}$ and $W_{q}^{(l)}$ are trainable parameters.

\subsection{Dual Relation-enhanced Loss} 

The dual relation-enhanced loss includes a support-level loss and a query-level loss to promote label knowledge propagation and  multi-label inference. 

\subsubsection{Support-level Loss} The support-level loss enhances the relation strength between instances with the same label connected to the support instance and weakens the relation strength between those with different labels.  
The ground-truth label relation of an instance pair is given as follows. 

\begin{equation}
y_{ij} = \left\{\begin{matrix} 
   1 , &\ if \ y_i = y_j, v_i \in \mathcal{S},  v_j \in \{\mathcal{S},\mathcal{Q}\}, \\
0, &\ if \ y_i \ne y_j, v_i \in \mathcal{S},  v_j \in \{\mathcal{S},\mathcal{Q}\},\\
\end{matrix}\right.
\label{initedge}
\end{equation}
where $y_i$ and $y_j$ are the labels of $v_i$ and $v_j$, respectively. $y_{ij} =1$ denotes that  $v_i$ and $v_j$ belong to the same class.

Any $i^{th}$ support instance is considered an anchor, where $v_i \in  \mathcal{S}$. 
The remaining instances are constructed a set as $J=\{\mathcal{S},\mathcal{Q}\} \backslash \{v_i\}$.
A positive set is $\Omega_{pos}^s = \{e_{ij}^{(l)}|y_{ij}=1, i \ne j \}$, and a negative set is $\Omega_{neg}^s = \{e_{ij}^{(l)}|y_{ij}=0, i \ne j \}$. 

\begin{equation}
\begin{aligned}
    \mathcal{L}_{s} =& \sum_{l=0}\frac{1}{|\mathcal{S}|} \sum_{v_i \in \mathcal{S}}(\log (1 + \sum\limits_{\substack{ e_{ij}^{(l)} \in \Omega_{neg}^s }} {\text{exp}({{e_{ij}^{(l)}}})} ) + \\
    &\log (1 + \sum\limits_{\substack{ e_{ij}^{(l)} \in \Omega_{pos}^s }} {\text{exp}({ - {e_{ij}^{(l)}}})} )),
\end{aligned}
\end{equation}

From the above formula, we can observe that the relation strength scores of connected instances derived from the same class are greater than value 0, while those with different classes are less than value 0. 
In this way, the support-level loss encourages the target node to learn information from similar neighbors and pushes it away from dissimilar neighbors to guide label knowledge propagation effectively.

\subsubsection{Query-level Loss}
The edge feature $e_{ij}$ is regarded as a probability that two connected nodes $v_i$ and $v_j$ are from the same class.
After $L$ layers of network propagation, the latest edge features from support nodes to query nodes are used for query inference. 
Specifically, each query node could be classified by final edge voting with support labels.
The prediction probability that a query node $v_i$ belongs to class $c_k$ could be formulated as $p_{i}^{c_k}$.

\begin{gather}
p_{i}^{c_k}=\sum_{v_j \in S} e_{ji}^{(l)}\delta (y^{s}_j=c_k), 
\end{gather}
where $y^s_j$ is the label of the support node $v_j$. $e_{ji}$ is the edge between the support node $v_j$ and the query node $v_i$. $\delta (y^s_j=c_k)$ is the Kronecker delta function that is equal to one when $y^s_j=c_k$ and zero otherwise.

Considering that a query may contain multiple user intents, the positive set for the $i^{th}$ query node is defined as 
$\Omega_{pos}^q  = \{ p_{i}^{c_k}|y_{i}^{q} = c_k\} $, which indicates the prediction probability set between a query node and its corresponding classes.
In contrast, a negative set is defined as $\Omega_{neg}^q  = \{ p_{i}^{c_k} |y_{i}^{q} \ne c_k\} $, which indicates the prediction probability set between a query node and its irrelevant classes. 
The training objective is used to minimize the following loss function.
\begin{equation}
\begin{aligned}
    \mathcal{L}_\textit{q} =& \frac{1}{|\mathcal{Q}|} \sum_{i = 1}^{|\mathcal{Q}|}(\log (1 + \sum\limits_{\substack{p_i^{c_k} \in \Omega_{neg}^q }} {{\text{exp}({p_{i}^{c_k}})}} ) + \\
    &\log (1 + \sum\limits_{\substack{p_{i}^{c_k} \in \Omega_{pos}^q }} {{\text{exp}( - {p_{i}^{c_k}})}} )),
\end{aligned}
\end{equation}

The aforementioned loss function aims to ensure that the scores in $\Omega^{q}_{pos}$ are greater than value $0$ and the scores in $\Omega^{q}_{neg}$ are less than value $0$. Therefore, the label $\hat{y_i}$ of the query instance is formulated as follows:

\begin{equation}
\hat{y_i}^{c_k} = \left\{\begin{matrix} 
   1 , &\ if \ p_{i}^{c_k} >0\\
0, &\ if \ p_{i}^{c_k} <0\\
\end{matrix}\right.
\end{equation}
where $\hat{y_i}^{c_k}=1$ shows that ${v}_i$ is predicted to class $c_k$, whereas  $\hat{y_i}^{c_k}=0$ shows that $v_i$ does not belong to $c_k$.

\subsection{Training Objective}
So far, we have introduced an instance relation learning network to effectively tackle the few-shot MID task.
The overall training objective is written as follows:
\begin{equation}
\mathcal{L} = \alpha \mathcal{L}_{s} + \beta  \mathcal{L}_{q},
\end{equation}
where $\alpha$ and $\beta$ are hyper-parameters.

\section{Experiments}

\subsection{Experimental Setup}
\paragraph{Datasets.}
We conduct experiments on the benchmark dataset TourSG \cite{williams2012dialog}. TourSG comprises six domains: Itinerary (It), Accommodation (Ac), Attraction (At), Food (Fo), Transportation (Tr), and Shopping (Sh).
The detailed statistics are presented in Table \ref{dataset}.

\paragraph{Evaluation Metric.}
Following previous works \cite{HouDetection,zhang2023dual}, we use AUC and Macro-F1 scores to evaluate the performance of our proposed method. 

\paragraph{Implementation Details.}

The proposed method is implemented with PyTorch (version 1.10.0) on a single GPU (RTX 3090 Ti) with CUDA version 11.3.
We use the pre-trained language model BERT-base \cite{devlin2019bert} as our encoder for $H$ (see Feature Extraction).
The AdamW optimizer trains the model with a learning rate of 5e-5.
Meanwhile, we use the GradualWarmupScheduler to optimize the learning rate and set the warmup proportion to 0.05.
The hyper-parameters $\alpha $ and $\beta $ are fixed as 0.1 and 1.
We set L to 2 as the number of cycles for the updates in the overall architecture. 
We test and validate the model on two domains, respectively, and train it on the remaining domains.
100 meta-tasks are randomly sampled for training, validation, and testing in every epoch.
We evaluate Macro-F1 scores in each meta-task and obtain the average F1 scores for all meta-tasks. Finally, the results of the test domain are reported when the validation domain obtains the best results.

\begin{table}[t]
\centering
\begin{tabular}{c cccccc ccc }
\toprule

\textbf{Dataset} & It & Ac & At & Fo & Tr & Sh  \\\cmidrule(lr){1-1} \cmidrule(lr){2-7}
 \#cla. & 15 & 17 & 18 & 18 & 17 & 16\\
 \#ins. & 397 & 1839 & 6162 & 2154 & 2493 & 1278   \\  \bottomrule
   
\end{tabular}
\caption{Statistics of datasets. \#cla. denotes the number of intents, and \#ins. indicates the number of utterances. }
	\label{dataset}
\end{table}

\subsection{Baselines}
We compare the proposed method with a series of strong baselines.
AWATT \cite{hu2021multi} designs a support-set attention mechanism and a query-set attention mechanism for learning representations of multi-label instances. Then, they use reinforcement learning to obtain a fixed threshold to identify multiple categories from these representations.
HATT \cite{gao2019hybrid} and {LDF} \cite{zhaoetal2022label}  use an empirical threshold to achieve multi-label results.
Besides, HATT utilizes instance-level and feature-level attention mechanisms to learn robust instance representations. 
{LDF} uses contrastive learning to learn different representations of multi-label instances. 
{CTLR} \cite{HouDetection} combines an instance and its label text to learn representations for multiple labels. Then, it leverages a meta-calibrated method to learn a dynamic threshold.
In addition, LPN \cite{liu2022label} and DCKPN~\cite{zhang2023dual} use label count estimation. They use an adaptive neural network to predict the label number of each instance. 
Besides, we also compare our method with Large Language Models (LLMs), such as Llama3-70b \cite{touvron2023llama} and GhatGLM4 \cite{glm2024chatglm}.

\begin{table*}[ht]
\centering
\begin{tabular}{l cc cc cc cc cc cc}
\toprule
\multicolumn{1}{c}{\multirow{2}{*}{\textbf{Models}}}& \multicolumn{2}{c}{It} & \multicolumn{2}{c}{Ac} &\multicolumn{2}{c}{At} &\multicolumn{2}{c}{Fo} &\multicolumn{2}{c}{Sh} &\multicolumn{2}{c}{Tr}  \\
\cmidrule(lr){2-3}\cmidrule(lr){4-5}\cmidrule(lr){6-7}\cmidrule(lr){8-9}\cmidrule(lr){10-11}\cmidrule(lr){12-13}
 & 1-shot & 3-shot  & 1-shot & 3-shot  & 1-shot & 3-shot& 1-shot & 3-shot & 1-shot & 3-shot & 1-shot & 3-shot\\\midrule
 \textbf{Llama3}  &  - &  -  &  - &  - &  -  &  -&  - &  -  &  -&  - &  -  &  -    \\
 \textbf{GhatGLM4}  &  - &  -  &  - &  - &  -  &  -&  - &  -  &  -&  - &  -  &  -   \\
 \textbf{CTLR} &  63.32 & 69.32 & 60.37 & 71.24 & 64.87 & 71.31 & 62.92 & 69.65 & 63.90 & 69.56 & 62.83 & 70.70       \\
 \textbf{HATT}  &    68.96 & 73.03 & 79.93 & 80.97 & 78.53 & 83.35 & 79.35 & 84.35 & 76.74 & 80.90 & 81.02 & 84.78       \\
 \textbf{LPN} &   65.89 & 73.51 & 73.36 & 79.24 & 74.71 & 82.00 & 69.04 & 79.27 & 70.54 & 78.58 & 70.58 & 81.17      \\
  \textbf{AWATT}  &  68.76 & 76.99 & 80.26 & 84.81 & 78.18 & 85.44 & 79.13 & 86.50 & 77.72 & 84.38 & 80.33 & 87.06    \\
 \textbf{LDF} &   68.46 & 75.88 & 80.46 & 84.17 & 80.38 & 85.18 & 79.50 & 86.47 & 77.08 & 84.49 & 79.83 & 87.83       \\
 \textbf{DCKPN} & 73.21 & 76.80 & 81.23 & 83.83 & 79.86 & 85.00 & 81.55 & 85.86 & 78.46 &  83.32  & 82.37 & 86.66\\
  \textbf{Ours} & \textbf{83.09} & \textbf{82.59} & \textbf{88.26} & \textbf{86.43} & \textbf{87.08} & \textbf{86.57} & \textbf{88.04} & \textbf{87.76} & \textbf{86.89} & \textbf{86.28} & \textbf{88.89} & \textbf{89.20}\\
\bottomrule
\end{tabular}
\caption{Comparison of AUC in 5-way scenarios.}
	\label{5w-AUC}
\end{table*}

\begin{table*}[!tbp]
\centering
\begin{tabular}{l cc cc cc cc cc cc}
\toprule
\multicolumn{1}{c}{\multirow{2}{*}{\textbf{Models}}}& \multicolumn{2}{c}{It} & \multicolumn{2}{c}{Ac} &\multicolumn{2}{c}{At} &\multicolumn{2}{c}{Fo} &\multicolumn{2}{c}{Sh} &\multicolumn{2}{c}{Tr}  \\
\cmidrule(lr){2-3}\cmidrule(lr){4-5}\cmidrule(lr){6-7}\cmidrule(lr){8-9}\cmidrule(lr){10-11}\cmidrule(lr){12-13}
 & 1-shot & 3-shot  & 1-shot & 3-shot  & 1-shot & 3-shot& 1-shot & 3-shot & 1-shot & 3-shot & 1-shot & 3-shot\\\midrule
 \textbf{Llama3}  &  48.91  &  48.45 &  54.23  &  49.57 & 53.07  &  54.35 &  57.54 &  52.56 &  53.66 & 51.25  &52.35  & 51.77   \\
 \textbf{GhatGLM4}  &  32.40 & 33.56  & 39.26 & 40.51 & 38.02 & 47.67 &  46.44 & 41.63  &  35.40 & 31.01 &  36.49 &  37.18\\
 \textbf{CTLR} &  39.31 & 40.30 & 38.00 & 40.77 & 37.11 & 39.59 & 37.88 & 40.14 & 38.03 & 40.22 & 38.44 & 40.37      \\
  \textbf{HATT}  &  39.27 & 46.01 & 50.47 & 55.40 & 28.28 & 58.03 & 39.33 & 60.69 & 43.92 & 56.05 & 43.50 & 62.89         \\
   \textbf{LPN} &   31.98 & 42.91 & 39.69 & 49.90 & 40.94 & 53.45 & 35.00 & 50.13 & 37.50 & 49.65 & 36.61 & 53.28        \\
  \textbf{AWATT}  &    40.25 & 50.61 & 53.52 & 62.22 & 50.88 & 62.49 & 52.24 & 64.62 & 50.61 & 61.95 & 54.90 & 66.47  \\
 \textbf{LDF} &  39.83 & 49.64 & 54.10 & 61.07 & 53.89 & 62.18 & 52.52 & 65.01 & 49.56 & 62.06 & 54.06 & 67.55        \\
  \textbf{DCKPN} & 45.73 & 53.35 & 57.89 & 62.37 & 55.13  & 64.52 &  58.39 & 66.74 & 53.87 &  62.48  & 59.95 & 68.73 \\
 \textbf{Ours} &   \textbf{62.94} & \textbf{61.82} & \textbf{69.15} & \textbf{65.87} & \textbf{65.20} & \textbf{65.29} & \textbf{68.78} & \textbf{68.24} & \textbf{64.98} & \textbf{65.31} & \textbf{71.24} & \textbf{71.80}  \\
\bottomrule
\end{tabular}
\caption{Comparison of Macro-F1 scores in 5-way scenarios.}
	\label{5w-F1}
\end{table*}

\begin{table*}[!tbp]
\centering
\begin{tabular}{l cc cc cc cc cc cc}
\toprule
\multicolumn{1}{c}{\multirow{2}{*}{\textbf{Models}}}& \multicolumn{2}{c}{It} & \multicolumn{2}{c}{Ac} &\multicolumn{2}{c}{At} &\multicolumn{2}{c}{Fo} &\multicolumn{2}{c}{Sh} &\multicolumn{2}{c}{Tr}  \\
\cmidrule(lr){2-3}\cmidrule(lr){4-5}\cmidrule(lr){6-7}\cmidrule(lr){8-9}\cmidrule(lr){10-11}\cmidrule(lr){12-13}
 & 1-shot & 3-shot  & 1-shot & 3-shot  & 1-shot & 3-shot& 1-shot & 3-shot & 1-shot & 3-shot & 1-shot & 3-shot\\\midrule
 \textbf{Llama3}  &  - &  -  &  - &  - &  -  &  -&  - &  -  &  -&  - &  -  &  -   \\
 \textbf{GhatGLM4}  &  - &  -  &  - &  - &  -  &  -&  - &  -  &  -&  - &  -  &  -  \\
 \textbf{CTLR} &   62.01 & 68.76 & 65.93 & 71.69 & 65.32 & 73.43 & 62.73 & 69.87 & 62.66 & 69.92 & 63.67 & 70.64     \\
  \textbf{HATT}  &   67.05 & 72.44 & 77.41 & 78.84 & 76.83 & 81.42 & 76.99 & 81.60 & 74.57 & 79.44 & 77.02 & 82.39        \\
   \textbf{LPN} &    65.40 & 74.09 & 72.90 & 79.18 & 73.70 & 81.59 & 68.94 & 79.50 & 69.17 & 78.21 & 69.52 & 80.55    \\
  \textbf{AWATT}  &  66.27 & 73.98 & 77.64 & 81.70 & 77.79 & 83.68 & 77.23 & 84.16 & 74.71 & 81.84 & 77.48 & 84.72  \\
 \textbf{LDF} &  66.58 & 76.14 & 77.86 & 81.46 & 77.79 & 83.96 & 77.39 & 84.00 & 75.23 & 81.53 & 77.14 & 84.83         \\
  \textbf{DCKPN} & 69.50 & 71.88 & 76.03 & 77.62 & 75.20 & 78.72 & 76.84 & 79.28 & 72.64 &  77.85  & 75.92 & 81.22 \\
 \textbf{Ours} &  \textbf{81.39} & \textbf{80.75} & \textbf{87.20} & \textbf{86.01} & \textbf{86.89} & \textbf{85.99} & \textbf{87.18} & \textbf{86.73} & \textbf{85.33} & \textbf{85.64} & \textbf{87.14} & \textbf{87.68}     \\
\bottomrule
\end{tabular}
\caption{Comparison of AUC in 10-way scenarios.}
	\label{10w-AUC}
\end{table*}

\begin{table*}[!tbp]
\centering
\begin{tabular}{l cc cc cc cc cc cc}
\toprule
\multicolumn{1}{c}{\multirow{2}{*}{\textbf{Models}}}& \multicolumn{2}{c}{It} & \multicolumn{2}{c}{Ac} &\multicolumn{2}{c}{At} &\multicolumn{2}{c}{Fo} &\multicolumn{2}{c}{Sh} &\multicolumn{2}{c}{Tr}  \\
\cmidrule(lr){2-3}\cmidrule(lr){4-5}\cmidrule(lr){6-7}\cmidrule(lr){8-9}\cmidrule(lr){10-11}\cmidrule(lr){12-13}
 & 1-shot & 3-shot  & 1-shot & 3-shot  & 1-shot & 3-shot& 1-shot & 3-shot & 1-shot & 3-shot & 1-shot & 3-shot\\\midrule
 \textbf{Llama3}  &  41.90   & 38.42  &  48.39 & 45.42  & 48.33  &  46.22 &  50.84 & 44.07  & 44.64 & 34.58 & 42.36  & 31.81\\
 \textbf{GhatGLM4}  & 26.62  & 29.66   & 34.95  & 38.00  &  35.60 & 37.53  &  31.97 & 37.40  &  30.25  & 28.36 & 32.11 & 25.79  \\
 \textbf{CTLR} &   26.46 & 28.72 & 26.06 & 28.41 & 24.56 & 28.13 & 24.72 & 27.27 & 25.45 & 27.34 & 25.96 & 27.91       \\
  \textbf{HATT}  &  21.58 & 36.39 & 30.25 & 45.26 & 36.74 & 47.34 & 25.30 & 48.08 & 29.81 & 44.55 & 32.66 & 50.52       \\
   \textbf{LPN} &  22.46 & 32.15 & 27.44 & 39.19 & 29.12 & 43.41 & 33.33 & 38.73 & 24.81 & 38.05 & 33.67 & 42.50       \\
  \textbf{AWATT}  & 28.93 & 37.20 & 41.85 & 49.51 & 41.01 & 51.32 & 40.99 & 52.25 & 39.16 & 48.81 & 43.62 & 55.17   \\
 \textbf{LDF} &    29.40 & 40.13 & 42.24 & \textbf{49.57} & 40.89 & \textbf{51.60} & 41.03 & 52.02 & 39.00 & \textbf{48.90} & 43.07 & 54.73    \\
  \textbf{DCKPN} & 34.20 & 39.18 & 41.93 & 48.00 & 39.29 & 48.83 & 46.58 & \textbf{53.31} & 38.48 &  48.57  & 39.66 & 55.00 \\
 \textbf{Ours} &   \textbf{47.45} & \textbf{47.76} & \textbf{51.34} & 47.48 & \textbf{49.00} & {47.10} & \textbf{51.55} & {52.99} & \textbf{48.43} & {47.74} & \textbf{55.34} & \textbf{55.65}    \\
\bottomrule
\end{tabular}
\caption{Comparison of Macro-F1 scores in 10-way scenarios.}
	\label{10w-F1}
\end{table*}

\subsection{Main Results}

We conduct extensive experiments and
report results in Tables \ref{5w-AUC}, \ref{5w-F1}, \ref{10w-AUC} and \ref{10w-F1}. The best results are highlighted in bold, with the following observations.

(1) Our proposed method achieves an average of 5.24\% AUC and 7.62\% Macro-F1 improvement in the 5-way scenario. In Table \ref{5w-AUC} and  Table \ref{5w-F1}, the proposed method improves upon the most competitive baseline DCKPN by 6.49\%-9.88\% AUC and 10.07\%-17.21\% Macro-F1 scores in the 5-way 1-shot setting, with an average improvement of 7.60\% AUC and 11.89\% Macro-F1 scores. 
In terms of the 5-way 3-shot setting, the proposed method improves upon the most competitive baseline DCKPN by 1.57\%-5.79\% AUC and 0.77\%-8.47\% Macro-F1 scores, with an average improvement of 2.89\% AUC and 3.36\% Macro-F1 scores.

(2) Furthermore, in Table \ref{10w-AUC} and Table \ref{10w-F1}, our proposed method achieves an average of 9.60\% AUC and 5.73\% Macro-F1 scores improvement in the 10-way scenario. 
In terms of the 10-way 1-shot setting, the proposed method improves upon the most competitive baseline DCKPN by 10.34\%-12.69\% AUC and 4.97\%-15.68\% Macro-F1 scores, with an average improvement of 11.50\% AUC and 10.49\% Macro-F1 scores. 
In terms of the 10-way 3-shot setting, the proposed method improves upon the most competitive baseline DCKPN by 6.46\%-8.87\% AUC, with an average improvement of 7.71\% AUC.
Compared to other methods, our 1-shot setting outperforms the 3-shot setting.
Our proposed method focuses on edges in the graph network, using these edges to directly indicate whether two instances belong to the same label. Therefore, each query node is classified based on final edge voting with support labels.
In 1-shot scenarios, fewer edges make optimization easier within the dual relation enhancement loss compared to in 3-shot scenarios, which is why 1-shot scenarios outperform 3-shot scenarios.
This advantage demonstrates that our proposed method can achieve better results with fewer samples.
However, DCKPN or LPN focuses on representation classification and ignores explicit interaction between instances, leading to low performance in 1-shot scenarios due to error propagation. 
Therefore, our proposed method guides the multi-label inference effectively. 


(3) Compared to LLMs, our proposed method consistently outperforms GhatGLM4 and Llama3. 
For few-shot MID, the task requires LLMs to extract all relevant intents from a given utterance. 
However, LLMs cannot understand dialogue diversity in different domains. 
Even with a few demonstration examples, LLMs may generate outputs that are not fully aligned with the expected labels. 
In the example ``\textit{What time is my meeting and what day?}", LLMs might only generate outputs like ``\textit{request time}", whereas the ground-truth labels are ``\textit{request time}" and ``\textit{request day}".
These incomplete outputs can degrade the model's performance, making it difficult for LLMs to handle multi-label tasks effectively.

\begin{table}[t]
\centering
\resizebox{1\columnwidth}{13mm}{

\begin{tabular}{lcccc}
\toprule
\textbf{Model} &\textbf{AUC}&\textbf{$\Delta $ AUC}& \textbf{F1}&\textbf{$\Delta $ F1}\\\midrule
Full model & \textbf{88.04}&  &  \textbf{68.78} &  \\
w/o label knowledge & 76.50 & -11.54  &  53.42   & -15.36  \\
w/o node pairwise logits &   86.39   & -1.65  &    66.84   & -1.94\\
w/o dual loss &  78.62  & -9.42  &  44.75   & -24.03  \\
w/o support-level loss &  85.87  &  -2.17 &  66.57   & -2.21 \\

\bottomrule
\end{tabular}
}
\caption{An ablation study on the 5-way 1-shot scenario.}
	\label{abla}
\end{table}

\subsection{Impact of Shot Number}
\begin{figure}[t]
 \centering
 \includegraphics[width=3.4 in, height=1.6 in]{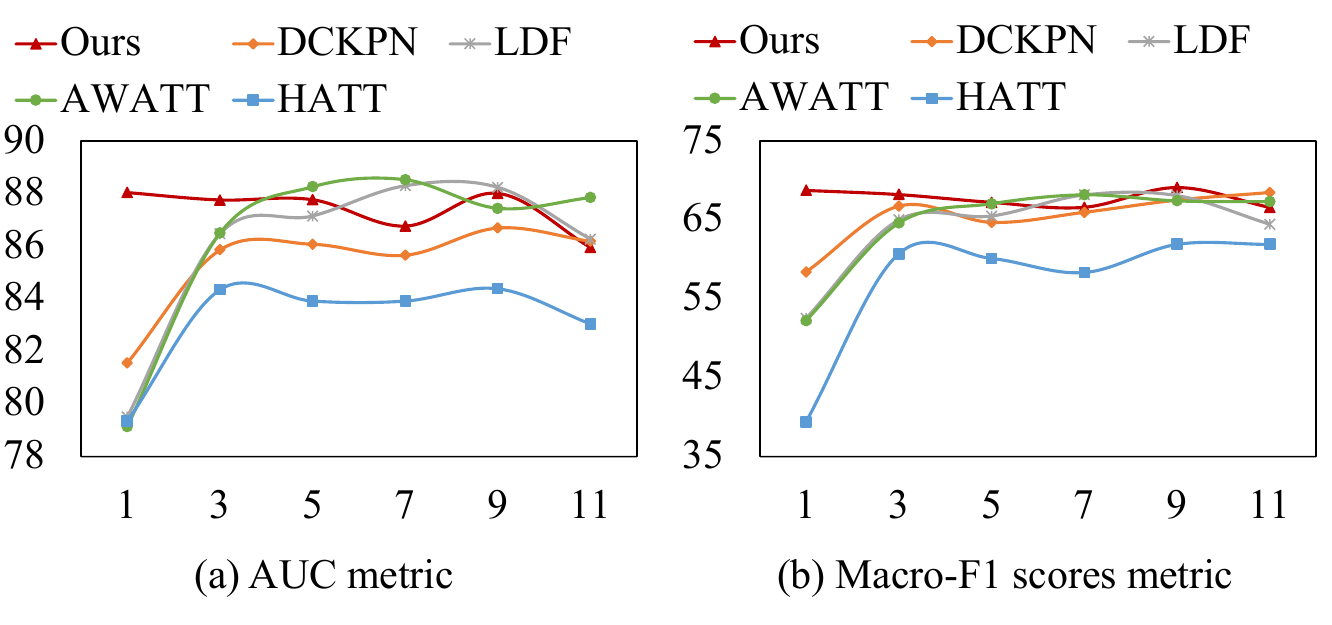}
 \caption{The impact of shot number.} 
 \label{qushi}  
\end{figure}

To investigate the impact of the shot number for each class, we evaluate our proposed method with one to eleven shots in the food domain. Figure \ref{qushi} shows the performances of methods in 5-way scenarios. 
The performances of DCKPN, LDF, AWATT, and HATT improve at first then fluctuate. The reason is that they focus on representation classification and rely on the number of labeled data to learn great instance representations.
However, these methods underperform in the 1-shot setting.
Our proposed method focuses on edges in the graph structure and utilizes the relation strength between instances to guide multi-label inference.
As discussed in the main results, optimizing the training objectives in fewer instances would be easy, which achieves great performance.
Our proposed method obtains the best results on the 1-shot setting, demonstrating the effectiveness of the model.

\subsection{Ablation Study}
In Table~\ref{abla}, we conduct an ablation study. 
(1) {``w/o label knowledge''.}  We remove class descriptions in the support set. 
The negative results suggest that class descriptions promote label knowledge propagation in the network. 
(2) {``w/o node pairwise logits''.}
We replace node pairwise logits with cosine similarity as edge features in the network. 
The low performance indicates that node pairwise logits play a positive effect by utilizing key and query of instances.
(3) {``w/o dual loss''.} After we remove the dual relation-enhanced loss, the performance drops a lot.
The negative results suggest that enhancing support-
and query-level relation strength is a key point for multi-label prediction.
(4)  {``w/o support-level loss''.} 
 Though we only remove the support-level loss, the performance is reduced. 
In conclusion, the complete model outperforms all ablation studies and achieves the best performance.

\section{Conclusion}

We propose a multi-label joint learning method for few-shot MID, addressing the negative effect caused by error propagation.
The proposed method explicitly models the relationship between instances and utilizes the relation strength between an instance pair to directly indicate whether these two instances belong to the same label, guiding multi-label inference effectively in an end-to-end manner. 
Besides, we design a dual relation-enhanced loss, which enhances interaction relations between support- and query-level instances and strengthens label knowledge propagation to further improve performance. 
Experiments show that we significantly outperform strong baselines, esp. the 1-shot setting.

\section{Limitation}
We incorporate class descriptions into the graph network to facilitate label knowledge propagation. Given that class descriptions are often brief and may provide limited context, their contribution to the overall model may be constrained. To improve performance, we plan to integrate more comprehensive external knowledge sources (e.g., domain-specific ontologies) into the graph construction, ultimately leading to better classification and prediction outcomes.
Besides, our proposed method focuses on the few-shot MID task in dialogue
systems. In the following work, we will explore other applications, such as health care.

\section*{Acknowledgements}
This research is supported by the National Science and Technology Major Project (No. 2021ZD0111202).

\bibliographystyle{named}
\bibliography{ijcai25}

\end{document}